%% file: bare_jrnl_new_sample4.tex
\documentclass[lettersize,journal]{IEEEtran}
\usepackage{amsmath,amsfonts}
\usepackage{algorithmic}
\usepackage{algorithm}
\usepackage{array}
\usepackage[caption=false,font=normalsize,labelfont=sf,textfont=sf]{subfig}
\usepackage{textcomp}
\usepackage{stfloats}
\usepackage{url}
\usepackage{verbatim}
\usepackage{graphicx}
\usepackage{cite}
\usepackage{multirow}
\usepackage{booktabs}
\usepackage{caption}
\usepackage{pifont}

\begin{document}

\title{Learning Downstream Task by Selectively Capturing Complementary  Knowledge from Multiple Self-supervisedly Learning Pretexts}

\author{
	\IEEEauthorblockN{
		Jiayu Yao \IEEEauthorrefmark{1}, 
		Qingyuan Wu \IEEEauthorrefmark{2}, 
		Quan Feng \IEEEauthorrefmark{3}, 
		Songcan Chen \IEEEauthorrefmark{4} \\
	}
	\IEEEauthorblockA{
		\IEEEauthorrefmark{1}
		Nanjing University of Aeronautics and Astronautics,
		Email: jiayu\_yao@nuaa.edu.cn
	}

	\IEEEauthorblockA{
		\IEEEauthorrefmark{2}
		Nanjing University of Aeronautics and Astronautics,
		Email: wuqingyuan@nuaa.edu.cn
	}

	\IEEEauthorblockA{
		\IEEEauthorrefmark{3}
		Nanjing University of Aeronautics and Astronautics,
		Email: fenquan@nuaa.edu.cn
	}

	\IEEEauthorblockA{
		\IEEEauthorrefmark{4}
		Nanjing University of Aeronautics and Astronautics,
		Email: s.chen@nuaa.edu.cn
	}
}


\markboth{IEEE TRANSACTIONS ON IMAGE PROCESSINGl}%
{Shell \MakeLowercase{\textit{et al.}}: A Sample Article Using IEEEtran.cls for IEEE Journals}


\maketitle

\begin{abstract}
 Self-supervised learning (SSL), as a newly emerging unsupervised representation learning paradigm, generally follows a two-stage learning pipeline: 1) learning invariant and discriminative representations with auto-annotation pretext(s), then 2) transferring the representations to assist downstream task(s). Such two stages are usually implemented separately, making the learned representation learned agnostic to the downstream tasks. Currently, most works are devoted to exploring the first stage. Whereas, it is less studied on how to learn downstream tasks with limited labeled data using the already learned representations. Especially, it is crucial and challenging to selectively utilize the complementary representations from diverse pretexts for a downstream task. In this paper, we technically propose a novel solution by leveraging the attention mechanism to adaptively squeeze suitable representations for the tasks. Meanwhile, resorting to information theory, we theoretically prove that gathering representation from diverse pretexts is more effective than a single one. Extensive experiments validate that our scheme significantly exceeds current popular pretext-matching based methods in gathering knowledge and relieving negative transfer in downstream tasks.                                                                                                                                                   
\end{abstract}


\section{Introduction}
\begin{figure}[h]
	\centering
	\includegraphics[width=8.5cm] {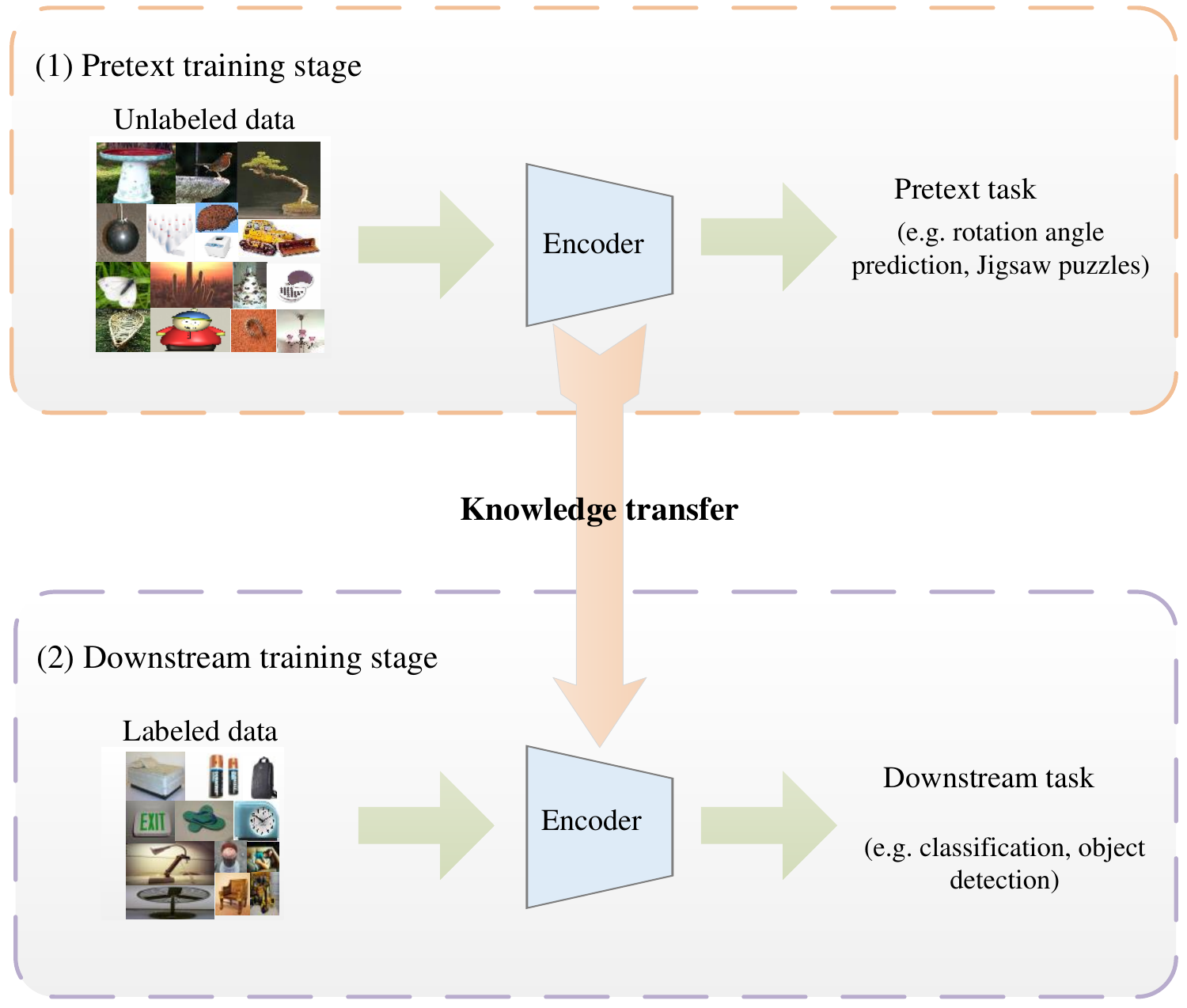}
	\caption{Self-supervised learning pipeline, (1) learning knowledge/representation with auto-annotation pretext task(s). (2) transferring knowledge to the specific downstream tasks as the end-goal.}
	\label{SSL_pipeline}
\end{figure}

\IEEEPARstart{T}{he} success of supervised learning as a learning paradigm benefits from large-scale labeled data \cite{wang2020revisiting}. In recent years, the progress of supervised learning has gradually been limited by the availability of large amounts of well-labeled data. In the real world, unlabeled data is easier to obtain \cite{qi2020small}. These data generally contain generalizable knowledge for learning new tasks. Nevertheless, learning discriminative representations directly from unlabeled data is a very challenging problem. Fortunately, SSL can overcome this problem through automatic annotation of the data's own attributes. As illustrated in Figure \ref{SSL_pipeline}, SSL usually follows a two stage pipeline: 1) learning the representation with auto-annotated pretext; 2) transferring the representation to the downstream tasks.

Most current popular works only focus on the first-stage, are dedicated to learning good representation \cite{gidaris2018unsupervised,robinson2020contrastive}. \emph{E.g.}, \cite{he2020momentum} the author learns invariant visual representations by contrasting negative pairs stored in a dynamic dictionary. \cite{chen2020simple} the author proposes to maximize the consistency between different augmented views of the same data sample in the latent space to obtain a more favorable visual representation. Excitingly, the performance of these methods compared with the performance of the supervised learning paradigm has gradually shown comparable results and even better in some specific scenarios.

By contrast, the understudied second-stage pipeline still needs further exploration. From the perspective of knowledge transfer, general transfer learning transfers knowledge from supervised pre-trained tasks to semantically similar target/downstream tasks. SSL typically transfers the representations learned under unsupervised pretext(s) to labeled downstream tasks. The semantics of the self-labeled and/or contrastive information used as a pretext in the SSL scenario is different from the supervision label information of downstream tasks. In addition, the semantics are agnostic of the pretexts to downstream tasks. Such a dilemma may lead to suboptimal or even unfavorable generalization performance for downstream tasks. To the best of our knowledge, a tiny minority of studies tried to focus on the problems derived from this semantic difference. \emph{E.g}., \cite{renggli2020model} the author defines multiple model pools for existing models to determine which models can be transferred to the current task for fine-tuning. \cite{shu2021zoo} the author proposes to adaptively aggregate the pre-trained parameters of channel alignment that can be learned by multiple source models to promote the transfer of multi-source knowledge to downstream tasks. The similar strategy of these methods is to pick the most similar semantic model from the pre-learned pretexts as the source of knowledge transfer for downstream tasks. However, this type of strategy ignores such problems. For downstream tasks, there may exist more complementary knowledge among different transfer sources. In practice, more likely, the knowledge which the downstream task needs distributes or scatters in (even semantically) different pretexts. Therefore, we need to select some pre-learned models for transfer purposes.

\begin{figure*}[t]
	\centering
	\includegraphics[width=1\textwidth]{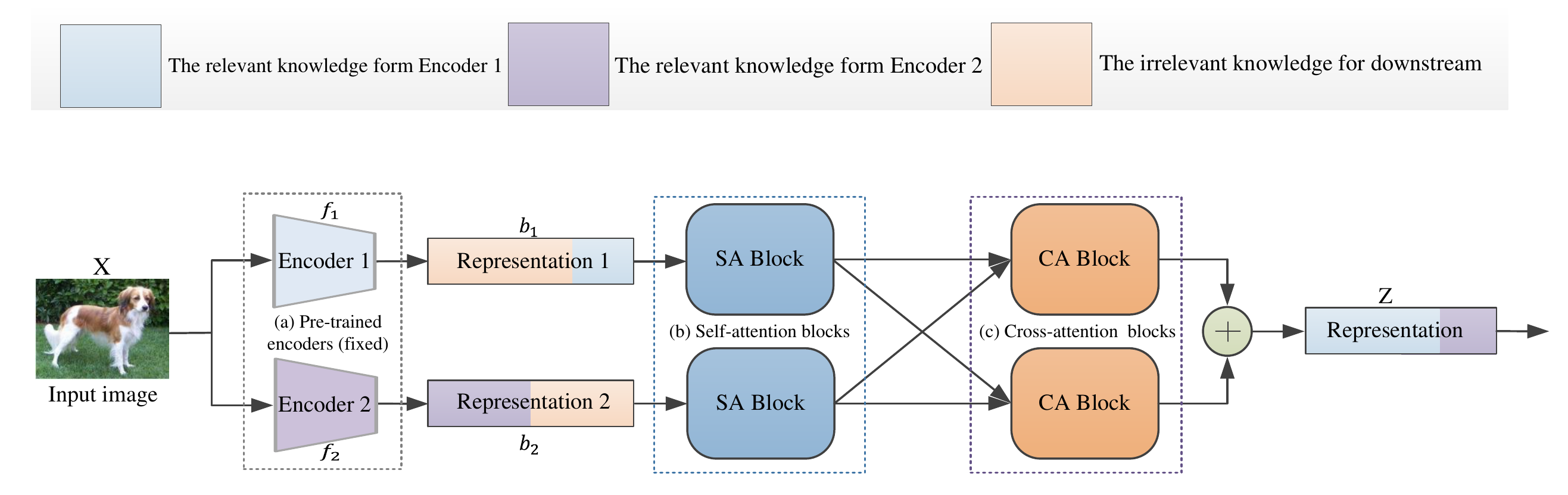} 
	\caption{Framework of our proposed scheme. First, we gather downstream task-specific knowledge with the self-attention block. Then,we align the complementary feature by a cross-attention block.
	}
	\label{fig:framework}
\end{figure*}

To this end, this paper focuses on how to squeeze out the favorable representations distributed in diverse learned pretexts. Specifically, we firstly theoretically analyze using the the information theory that diverse pretexts can provide more favorable complementary knowledge to the downstream tasks, $i.e.$, squeezing out helpful representations from the constructed representation bank will lead to much more mutual information associated with the downstream tasks to be learned. Next,  to guide purposeful squeeze, we propose an attention-based method that extracts diverse knowledge from the constructed representation bank and selectively focuses on the pivotal knowledge beneficial  for downstream tasks. Finally, extensive experiment results demonstrate that our method is significantly more competitive compared with state-of-the-art (SOTA).


We summarize our main contributions as follows: 
\begin{itemize}
\item It's the first time investigating relationships among diverse representations from multiple pre-trained SSL models. 
\item A beneficial conclusion that purposefully and strategically squeezing out diverse representational knowledge from the constructed representation bank is conductive for downstream tasks learning.

\item A novel attention-based framework that adaptively extracts the must representations from the constructed representations bank without matching the pretexts with downstream tasks.

\end{itemize}

The rest of this paper is organized as follows: Section 2 reviews some related work; Section 3 introduces our attention-based contrast learning method; Section 4 reports the experimental results, followed by conclusion in Section 5.

\section{Related Works}
\subsection{Self-supervised learning (SSL)}
Self-supervised learning is a rapidly developing new unsupervised representation learning paradigm, its principal merit is learning representation/knowledge without manual annotations. This paradigm generates supervision signals from intrinsic attributes or regularity of data ~\cite{gidaris2018unsupervised,caron2018deep}.
Currently, the  springing up contrast-based family (like MoCo \cite{he2020momentum}, SimCLR \cite{chen2020simple}, BYOL \cite{grill2020bootstrap}) even outperforms supervised counterparts  in some particular datasets or downstream tasks.

However, the conclusion in \cite{ericsson2020well} shows that there does not exist a best self-supervised representation over all tasks and just those downstream tasks similar to pretext will be generalized well, vice versa.
In practice, pretext semantics tends to be agnostic to downstream tasks.
Thus seeking a most similar pretext to downstream task ~\cite{nguyen2020leep,renggli2020model} or aggregating  representations of diverse pretexts ~\cite{ngiam2011multimodal,srivastava2014multimodal} becomes a key and has given rise to some methods.
The former either estimates the performance on downstream by Log Expected Empirical Prediction or fine-tunes the pre-trained models on downstream task,then transfers the best model to downstream tasks.
Doing so tends to overlook the complementarity from other pretexts, especially when the best one just possesses partial knowledge for the target task.
While fine-tuning on the model architecture also limits flexibility of the downstream models \cite{noroozi2018boosting}.
And even worse, the details, such as the architecture or weights, of the pre-learned models are all unavailable due to the privacy policy.
The latter simply aggregate representations as downstream task-specific representation from all pretexts without selection, which may invoke too much irrelevant information, leading to poor performance of downstream task \cite{tishby2015deep}.
Similar to the aggregation methods, but a major difference from them is that ours strategically squeezes out must representations for downstream task from the knowledge bank by the attention mechanism without the need of both pre-learned models' details and searching for single best pretext.

\subsection{Attention mechanism}
The attention mechanism is one of the important concepts in the current deep learning field. The attention mechanism is inspired by human biological systems that are more inclined to focus on specific areas of the object when processing information. Therefore, the attention mechanism aims to imitate human visual attention to capture pivotal object-related information. According to the realization of the attention mechanism, it can be briefly divided into the following categories:

1) $Hard ~attention$, which uses one-hot coding to obtain information about a specific area in the hidden sampling process of the coding layer. \emph{E.g}., \cite{papadopoulos2021hard} the author proposes a top-down method of traversing the image scale space, which only accesses the multi-scale hard attention network of the image area with the most information along the way for image classification. \cite{wang2020hard} the author designs an end-to-end hard attention network with encoders and decoders for retinal vessel segmentation. \cite{shankar2018surprisingly} the author proposes a joint hard attention model that couples the input state to the output state separately for natural language processing. However, this type of method requires Monte Carlo sampling or similar methods to estimate the gradient.

2) $Soft ~attention$, which uses the $[0,1]$ weight probability distribution to focus on the information of a specific area in the coding process of hidden layer sampling. \emph{E.g}., \cite{jain2020sarcasm} the author proposes an attention model combining bidirectional long-term and short-term memory, soft maximum attention layer and convolutional neural network for real-time sarcasm recognition. \cite{mcclenny2020self} the author designs a deep network that uses trainable weights as soft multiplication masks and trains adaptive weights and network weights at the same time. This type of method solves the problem of hard attention not being differentiable, and makes it utilize the forward propagation and backward feedback of the neural network to learn the attention weight.

3) \emph{Self-attention}, an attention variant, focuses more on internal correlations in data features. \emph{E.g}., \cite{guan2021predicting} the author designs a VisText self-attention module that combines relevant multi-modal features and clinical features to predict the risk of esophageal fistula. \cite{guo2021ssan} the author designs a separable self attention  module that sequentially models spatial and temporal correlations for video representation learning. This type of method reduces dependence on external information, thereby improving the reasoning time of the model.

4) \emph{Multi-headed attention}, which performs parallel computing on multiple attentions. \emph{E.g}., \cite{li2021cross} the author proposes an unsupervised multi-head attention network for cross-media hash retrieval of pictures and texts of semantic information. \cite{wang2020cascade} the author proposes a cascaded multi-head attention network composed of multi-head local self-attention and feature relations for action recognition. The advantage of this type of method is that it can pay attention to the information of multiple subspaces at the same time.

5) $Local ~attention$, which only focuses attention to the context information of the small window (the small area), but does not focus attention on the state of all coding hidden layers. \emph{E.g}., \cite{shang2021span} the author proposes a model that uses supervised dynamic local attention to explicitly capture structural information for continuous sentence classification. \cite{hanunggul2019impact} the author proposes to compare the local attention in the long short-term memory (LSTM) model to generate abstract text summarization. The characteristic of this kind of method is that more cost can be used to obtain miniaturized information on some key objectives.

6) $Global ~attention$, which takes into account both the object decoding hidden state and all encoded hidden states in the calculation process. \emph{E.g}., \cite{deng2021superpoint} the author proposes a global attention network composed of a point-independent global attention module and a point-dependent global attention module for point cloud semantic segmentation. \cite{ren2020scga} the author proposes an order sensitive global attention network for text and image feature enhancement. This type of method is computationally costly and easy to distract when dealing with long text sequences. 

Since this paper focus is how to use the existent attention mechanism to adaptively squeeze out the complementary knowledge from the bank, rather than itself. Therefore, just reviewing a few closely-related works as above.


\section{Method}
In practice, the semantics of the pre-learned pretexts are unnecessarily sufficiently relevant with the supervised information of the downstream task, leading to \textit{a semantic gap between pretext(s) and specific downstream task to great extent so that knowledge required by learning downstream task is NOT sufficient or deficient.}
Therefore, squeezing out knowledge from only one single pretext for a specific task very likely gives rise to a serious knowledge deficiency.
In this section, our work is to compensate for the deficiency by supplying the specific task with its must complementary knowledge from diverse learned pretexts.
Specifically, we propose an attention-based framework (Figure 2) to selectively extract or collect the representations distributed in diverse pretexts instead of searching for the single best pretext. The more details are described in the following subsections.
\subsection{Preliminaries}\label{preliminaries}
Given a set of $N$ pre-learned models $\mathcal{F} = \{f_{1},...,f_{N}\}$ via some contrastive learning such as MoCo \cite{he2020momentum}. For them we need not know their details such as data used in pretext training and the architecture of the model except for its penultimate layer. We aim to squeeze out knowledge from $\mathcal{F}$ to assist learning a downstream task $\mathcal{T}=\{(x,y)\}$.
With the $\mathcal{F}$, we get $x$ representation bank $\mathcal{B} = \{ b_i |b_i=f_{i}(x), i=1,...,N \}$, then we squeeze out so-needed representation $z$ from $\mathcal{B}$ for the task $\mathcal{T}$ by aggregating $z=agg(\mathcal{B})$.

An ideal representation $z^*$ should contain the maximal task-relevant knowledge assosiate with the specific downstream task $\mathcal{T}$ (maximized information) and minimal task-irrelevant knowledge (minimized irrelevant information) \cite{achille2018emergence,tishby2015deep}. Where ‘ideal representation’ follows \cite{tishby2000information}, it satisfies Definition 1.

\textbf{Definition 1}
	An ideal representation $z^*$ of $x$ should have the property $i.e.$,
	
	Sufficiency and Compactness:
\begin{equation}
z^{*}=\arg \min _{z^{s u f}}\left\{I\left(x ; z^{s u f}\right) \mid z^{s u f}=\arg \max _{z} I(y ; z)\right\},
\end{equation}
where $z^{suf}$ contains all information (maximal task-relevant knowledge) $y$ shares with input $x$. And $z^*$ is the most compact (minimal task-irrelevant knowledge) $z^{suf}$, $I(x,y)$ is the mutual information between $x$ and $y$.

Definition 1 implies Eq.(\ref{data-processing inequality}), which is so-called Data Processing Inequality (DPI).
\begin{equation}\label{data-processing inequality}
I(y;x)=I(y;z^{suf})=I(y;z^{*})\geq I(y;z).
\end{equation}

\begin{figure}[t]
	\centering
	\includegraphics[width=0.35\textwidth]{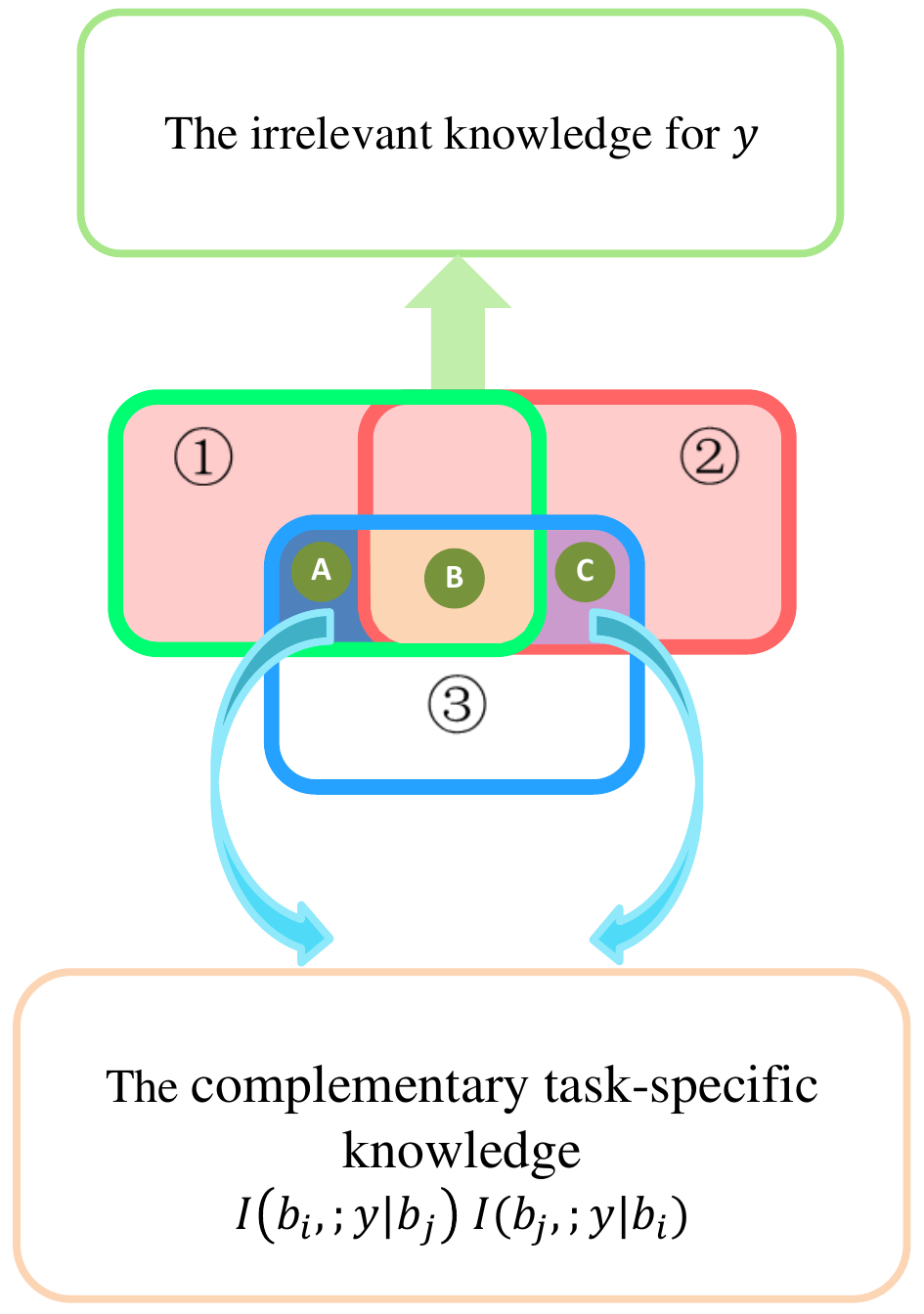}
	\caption{The illustration of complementary knowledge using the information diagram. \textcircled{1} and \textcircled{2}  represent pretext 1 and pretext 2, respectively, and \textcircled{3} represent downstream tasks. \textcircled{A}  and \textcircled{C} are complementary knowledge; \textcircled{B}  is the knowledge downstream task shares with both pretext 1 and pretext 2.}
	\label{fig:framework}
\end{figure}
 
\subsection{Complementarity of diverse models}\label{complementarity}
In practice, pretexts tend to be diverse, leading to diverse pre-learned models with different performances on a specific downstream task.
Such differences imply that diverse semantic knowledge available for the task.
Furthermore, we assume the available knowledge is complementary for a specific downstream task (Assumption 1) and corresponding information diagiam is shown in Figure 3.

\textbf{Assumption 1.}
Task-specific complementary knowledge: for any two different$(i\neq j)$ pre-trained models $f_i$ and $f_j$, the knowledge they provide for a specific downstream task $(x,y)$ is complementary, meaning Eq.(3),
	where $b_i=f_i(x)$ and $b_j=f_j(x)$.

\begin{equation}
I(b_i;y|b_j)> 0,
\end{equation}

For the pre-trained model $f_{i}$, we squeeze out downstream task-specific representation $\hat{z}_{i}$ from $b_i$.
According to Eq.(\ref{data-processing inequality}), the $\hat{z}_{i}$ has the property as following:
\begin{equation}\label{element-wise dpi}
\begin{array}{rr}
I(y;b_i)=I(y;\hat{z}_{i}^{suf})=I(y;\hat{z}_{i}^{*})\geq I(y;\hat{z}_{i}).
\end{array}
\end{equation}

Based on Assumption 1, for a specific downstream task $(x,y)$, the aggregated representation $z$ from $\mathcal{B}$ will be more sufficient than the $\hat{z}_{i}$ from a single pre-trained model as shown in Theorem 1:

\textbf{Theorem 1.} 	
When $z=agg(\mathcal{B})$ is the sufficient representation of $(b_1,b_2,...,b_N)$ for $y$, then
\begin{equation}
I(y ; z)>\max \left\{I\left(y ; \hat{z}_{1}\right), I\left(y ; \hat{z}_{x}\right), \ldots, I\left(y ; \hat{z}_{N}\right)\right\},
\end{equation}

$Proof$. Based on DPI, we get following mutual information inequality of $z$ and $y$:
\begin{equation}
I(y;z) \leq I(y;b_1,b_2,...,b_N),
\end{equation}

And Eq.(\ref{data-processing inequality}) implies that $z^{suf}$ is the sufficient representation of $(b_1,b_2,...,b_N)$ for $y$, thus
	\begin{equation}
	I(y;z^{suf}) = I(y;b_1,b_2,...,b_N).
	\end{equation}

So for any $b_i(1\leq i\leq N)$, based on the chain rule of mutual information, we have
	\begin{equation}
	\begin{array}{ll}
	I(y;z^{suf}) &= I(y;b_1,b_2,...,b_N)\\
	&= I(y;b_i) + I(b_1,...,b_{i-1},b_{i+1},...,b_{N};y|b_i),
	\end{array}
	\end{equation}

	Again, from the Eq.(3) in Assumption 1, we get
	\begin{equation}
	I(b_1,...,b_{i-1},b_{i+1},...,b_{N};y|b_i)>0,
	\end{equation}
	and
	\begin{equation}\label{conclusion equation}
	I(y;z^{suf}) > I(y;b_i),
	\end{equation}

Incorporating the Eqs.(\ref{element-wise dpi}), (6), (7), (\ref{conclusion equation}), we get
	\begin{equation}
	I(y;z^{suf}) > I(y;b_i) = I(y;\hat{z}_i^{suf}) \geq I(y;\hat{z}_i),
	\end{equation}
	
	Hence,
	\begin{equation}
	I(y;z^{suf}) > \max\{I(y;\hat{z}_1),I(y;\hat{z}_2),...,I(y;\hat{z}_N)\}.
	\end{equation}

\textbf{Remark:} Eq.(5) shows that \textit{an ideal reperesentation $z$ squeezed from bank $\mathcal{B}$ will contain more downstream knowledge than the $\hat{z}_{bm}$} learned from the best matching pretext ~\cite{nguyen2020leep,renggli2020model}, where
\begin{equation}
\hat{z}_{bm}=\arg \max \limits_{\hat{z}_i} \{I(y;\hat{z}_1),...,I(y;\hat{z}_i),...,I(y;\hat{z}_N)\}.
\end{equation}

\subsection{Self-Attention block \& Cross-Attention block}
In general, directly aggregating (e.g. adding and concatenating \cite{ngiam2011multimodal,srivastava2014multimodal}) all representations of the bank $\mathcal{B}$ will unfavorably introduce relatively much irrelevant information, resulting in poor performance on downstream task.
To alleviate the negative effect, we would like to squeeze out the must representation with both maximal task-relevant knowledge and minimal task-irrelevant knowledge as described in Definition 1.
The existent attention mechanism \cite{vaswani2017attention,hu2018squeeze} is able to selectively focus on pivotal information beneficial to the tasks and decrease the weights of irrelevant information.
Such unique merit offers a practical strategy for adaptively squeezing out desired representation from $\mathcal{B}$.
In this paper, we extract complementary knowledge from the representation bank $\mathcal{B}$ by two attention-based blocks as shown in Figure 4.
\subsubsection{Self-Attention block}
Inspired by \cite{hu2018squeeze}, we adaptiely select the task related knowledge within each pretext by Self-Attention Block (SAB) whose details are shown in Figure 4 (a).
We transform the $b_i \in \mathbb{R}^{1 \times d}$ into two different $query$ and $key$ to calculate the self-attention weight $w_{sa}(b_i)$ by Eq.(14), where $query(b_i)=q(b_i)=b_{i}W_{q}, key(b_i)=k(b_i)=b_{i}W_{k}$ and $W_{q},W_{k} \in \mathbb{R}^{d \times d}$.

\begin{equation}
w_{sa}(b_i) = sigmoid(q(b_i)k(b_i)^{T}),
\end{equation}
We further multiply $w_{sa}$ by the $value(b_i)=v(b_i)=b_{i}W_{v}$ ($W_{v} \in \mathbb{R}^{d \times d}$.) and calculate the final output as Eq.(15).
\begin{equation}
z^{self}_{i} = b_{i} + w_{sa}(b_i)v(b_i).
\end{equation}

\begin{figure*}[t!]
	\centering
	\includegraphics[width=0.75\textwidth]{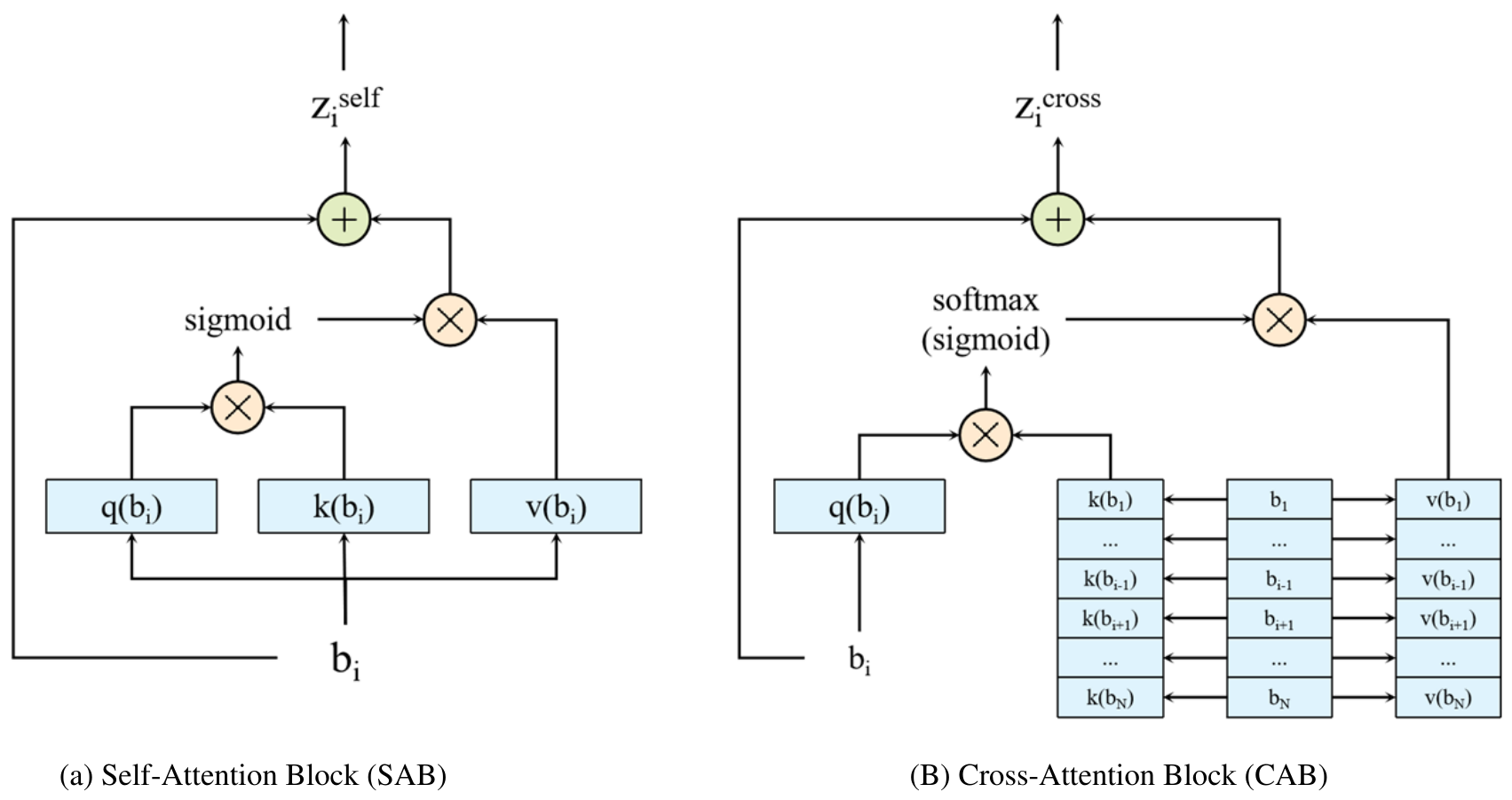}
	\caption{Illustrate two attention blocks in detail. (a) calculates the self-attention within pretext to selectively squeeze out task-relevant knowledge. (b) aggregates complementary knowledge among different pretexts.}
	\label{fig:framework}
\end{figure*}

\subsubsection{Cross-Attention block}
To selectively squeeze out the compact and complementary knowledge, we also construct Cross-Attention Block (CAB) similar to guided-attention layer \cite{yu2019deep} (Figure 4 (b)).
Given $N$ pretexts, as before, we get representation $b_i \in \mathbb{R}^{1 \times d} (i=1,..,N)$.
We obtain the attention weight(s) ($w_{ca}(b_i,b_j)$) by Eq.(16) measuring the similarity between $b_i$ and $b_j$ ($i \neq j$), where $q(b_i)=b_{i}W_{q}^{i}$, $k(b_i)= x_{i}W_{k}^{i}$ and $v(b_i)=b_{i}W_{v}^{i}$, $W_{q}^{i},W_{k}^{i},W_{v}^{i} \in \mathbb{R}^{d \times d}$

\begin{equation}
w_{ca}(b_i,b_j) = \left \{
\begin{array}{ll}
sigmoid(q(b_i)k(b_j)^{T}), & N=2, i \neq j\\
\frac{exp(q(b_i)k(b_j)^{T})}{\sum\limits_{j=1,j\neq i}exp(q(b_i)k(b_j)^{T})}, & N>2, i \neq j
\end{array}
\right.
\end{equation}
Accordingly, the output of the CAB is generated by Eq.(\ref{CA output}).
\begin{equation}
\label{CA output}
z_i^{cross}=b_i + \sum_{j=1,j\neq i}^{n}w_{ca}(b_i, b_j)v(b_j).
\end{equation}

\subsection{Architecture}
We adaptively squeeze out the representations from the bank for the downstream task by the architecture equipped with SAB and CAB as Figure 2.
We get the representation bank $\mathcal{B}=\{ b_{1},b_{2}\}$ of $x$ by $\mathcal{F}=\{ f_{1},f_{2}\}$.
For $b_1$ and $b_2$, we (1) firstly enlarge their own task-related knowledge by SA Blocks, and then (2) extract the complementary knowledge from the another by CA Blocks.
The logits are obtained by a fully-connected (FC) layer for downstream classification task.

\begin{table}[ht!]
\scriptsize
\centering
\caption{Characteristics of the experimental datasets.}
\scriptsize
	\label{exp:1}
\renewcommand\arraystretch{2.0}
	\setlength{\tabcolsep}{2.5mm}{
\begin{tabular}{c|c|c|c|c}
\bottomrule
\hline  Data set & Train & Test & Features & classes \\
\hline CIFAR10 & 50,000 & 10,000 & 1024 & 10 \\
\hline CIFAR100 & 50,000 & 10,000 & 1024 & 100 \\
\hline SVHN & 73,257 & 26,032 & 1024 & 10 \\
\hline TinyImagenet & 100,000 & 20,000 & 4,096 & 200 \\
\hline Chars74k & 7,705 & 4,348 & 1024 & 64 \\
\bottomrule
\end{tabular}}
\end{table}

\section{Experiment}\label{Experiment}
To further verify the effectiveness of the method proposed in Section 3, we conduct extensive image classification downstream tasks on benchmark datasets.

\subsection{Experimental Settings}
\subsubsection{Datasets}
We adopt four datasets, CIFAR10 \cite{krizhevsky2009learning}, CIFAR100 \cite{krizhevsky2009learning}, SVHN \cite{netzer2011reading} and TinyImagenet \cite{le2015tiny}, for pretext training.
And CIFAR10, CIFAR100, SVHN and Chars74k \cite{de2009character} are used as different four downstream task datasets. CIFAR10 has a total of 60000 color images from 10 natural image classes, with 6000 images per class. Each image size is 32 $\times$ 32, and the training and test sets are respectively 50000 and 10000. CIFAR100 is like CIFAR10, and it has 100 classes containing 600 images each. SVHN is a real world digit datasets containing 10 classes. It has 73257 images for training and 26032 digits for testing, and size is also like 32 $\times$ 32. TinyImagenet has a total of 200 classes with 500 images for each class for training, which is drawn from Imagenet  ILSVRC 2012 and down-sampled to 64 $\times$ 64. Chars74k, we only use Englishlmg subset, has 62 classes of letters and numbers, with 7705 images training set and 4348 images testing set, and images are resized to like 32 $\times$ 32. Table-I reports detailed information on datasets.
\begin{table*}[ht!]
\scriptsize
\centering
\caption{Classification accuracy (\%) of  transferting representation from single pre-trained model to different downstream tasks. The best result is indicated as bold.}
\scriptsize
	\label{exp:1}
\renewcommand\arraystretch{1.5}
	\setlength{\tabcolsep}{3.1mm}{
\begin{tabular}{c|c|cccc|ccc}
  \bottomrule
  \hline \multirow{2}{*}{ pre-trained model } & \multirow{2}{*}{ SA Block } & \multicolumn{4}{c|}{ downstream (100\% images and labels) } & \multicolumn{3}{c}{ downstream (10\% images and labels)} \\\cline { 3 - 9 } & & CIFAR10 & CIFAR100 & SVHN & Chars74K & CIFAR10 & CIFAR100 & SVHN \\\hline

  \multirow{2}{*}{CIFAR10}& w/o SA & 82.40& 46.42& 60.62&31.99             &$79.95$&$40.01$&$54.65$\\
  & w/ SA& \textbf{85.78}&\textbf{57.49}&\textbf{72.47}& \textbf{44.54}    &$\mathbf{80.57}$&$\mathbf{41.73}$&$\mathbf{66.87}$ \\ \hline
  \multirow{2}{*}{CIFAR100}&w/o SA& 75.64& 51.90&65.98&35.37               &$72.25$&$42.75$&$60.70$\\
  & w/ SA&\textbf{81.54}&\textbf{60.61}&\textbf{76.33}&\textbf{45.62}      &$\mathbf{74.62}$&$\mathbf{45.31}$&$\mathbf{70.87}$\\ \hline
  \multirow{2}{*}{Tiny}& w/o SA &72.98&43.36&60.84&30.58                   &$69.23$ & $34.77$ & $53.67$\\
  & w/ SA & \textbf{79.60}&\textbf{57.71}&\textbf{73.46}&\textbf{50.27}    &$\mathbf{74.16}$&$\mathbf{42.05}$&$\mathbf{68.48}$\\ \hline
  \multirow{2}{*}{SVHN}& w/o SA & 62.90& 32.90& 92.37 & 40.75              &$58.34$&$25.80$&$90.90$\\
  & w/ SA & \textbf{71.62}& \textbf{46.82}& \textbf{94.40}&\textbf{44.83}  &$\mathbf{66.08}$&$\mathbf{33.41}$&$\mathbf{91.27}$\\ \bottomrule
 \end{tabular}}
\label{single pre-trained model}
\end{table*}

\begin{table*}[ht!]
\scriptsize
\centering
\caption{Classification accuracy (\%) of different aggregation feature scheme, add, concatenate and our attention-based. The best result is indicated as bold and the second is underlined.}
\scriptsize
	\label{exp:1}
\renewcommand\arraystretch{1.2}
	\setlength{\tabcolsep}{2.0mm}{
\begin{tabular}{ccccccc}
\bottomrule
\multicolumn{1}{c|}{pre-trained models}   & CIFAR10+CIFAR100 & CIFAR10+Tiny & CIFAR100+Tiny & CIFAR10+SVHN & CIFAR100+SVHN & Tiny+SVHN   \\ \toprule
\multicolumn{7}{l}{CIFAR10}                                                                                                              \\ \hline
\multicolumn{1}{c|}{\textbf{add}}& 83.23& 83.15& 77.65& 82.71& 77.03& 73.57\\
\multicolumn{1}{c|}{\textbf{concatenate}}& 83.60& 83.38&78.34&83.01&77.48&74.41\\
\multicolumn{1}{c|}{\textbf{ours}}&\underline{86.08}&\underline{86.72}&\underline{85.28}&\underline{86.00}&\underline{83.21}&\underline{82.53} \\ \cline{2-7}
\multicolumn{1}{c|}{\textbf{supservised}} & \multicolumn{6}{c}{\textbf{95.32}}\\ \toprule[1pt]
\multicolumn{7}{l}{SVHN}                                                                                                                 \\ \hline
\multicolumn{1}{c|}{\textbf{add}}& 68.62&65.71&68.83&92.37&92.27&92.37\\
\multicolumn{1}{c|}{\textbf{concatenate}}&69.48&66.84& 69.38&92.52& 92.65& 92.49\\
\multicolumn{1}{c|}{\textbf{ours}}&\underline{80.61}&\underline{80.22}&\underline{81.18}&\underline{94.26}&\underline{94.66}&\underline{94.37} \\ \cline{2-7}
\multicolumn{1}{c|}{\textbf{supservised}} & \multicolumn{6}{c}{\textbf{96.55}} \\ \toprule
\multicolumn{7}{l}{CIFAR100} \\ \hline
\multicolumn{1}{c|}{\textbf{add}}& 54.9& 50.8& 54.03& 49.15& 53.23& 44.91\\
\multicolumn{1}{c|}{\textbf{concatenate}}& 55.01& 51.27& 54.3& 49.43 & 53.32& 45.36\\
\multicolumn{1}{c|}{\textbf{ours}}& \underline{62.04}& \underline{62.26}& \underline{62.26}& \underline{60.21}&\underline{61.16}& \underline{58.72} \\ \cline{2-7}
\multicolumn{1}{c|}{\textbf{supservised}} & \multicolumn{6}{c}{\textbf{77.35}}\\ \toprule
\multicolumn{7}{l}{Chars74k}                                                                                                             \\ \hline
\multicolumn{1}{c|}{\textbf{add}}&38.70& 36.99& 38.06& 43.64& 44.23& 44.27\\
\multicolumn{1}{c|}{\textbf{concatenate}}& 38.80& 36.81& 38.60& 44.27& 44.69& 43.79\\
\multicolumn{1}{c|}{\textbf{ours}}& \underline{47.56}& \underline{48.79}& \underline{48.29}&\underline{50.5}&\underline{51.29}&\underline{51.63} \\ \cline{2-7}
\multicolumn{1}{c|}{\textbf{supservised}} & \multicolumn{6}{c}{\textbf{66.8}}                                                           \\ \toprule[1pt]
\end{tabular}}
\label{444}
\end{table*}

\begin{table*}[ht!]
\scriptsize
\centering
\caption{Classification accurency (\%) of different aggregated shemes on 10\% of the labeled data. The best result is indicated as bold and the second is underlined.}
\scriptsize
	\label{exp:1}
\renewcommand\arraystretch{1.2}
	\setlength{\tabcolsep}{2.0mm}{
\begin{tabular}{ccccccc}
\bottomrule
\multicolumn{1}{c|}{pre-trained models}   & CIFAR10+CIFAR100& \multicolumn{1}{c}{CIFAR10+Tiny}& \multicolumn{1}{c}{CIFAR100+Tiny}&
\multicolumn{1}{c}{CIFAR10+SVHN}&\multicolumn{1}{c}{CIFAR100+SVHN}&\multicolumn{1}{c}{Tiny+SVHN}\\ \toprule
\multicolumn{7}{l}{10\% CIFAR10}                                                                                                                                                                                                                                       \\ \hline
\multicolumn{1}{c|}{\textbf{add}}&\multicolumn{1}{c}{\underline{80.59}}& \underline{80.42}& 73.75& 80.19 & \underline{73.23}& 69.34\\
\multicolumn{1}{c|}{\textbf{concatenate}} & \multicolumn{1}{c}{\textbf{80.72}} & 80.41 & \underline{74.04} & \underline{80.36}& 73.19& \underline{69.41}\\
\multicolumn{1}{c|}{\textbf{ours}}        & \multicolumn{1}{c}{80.53}&\textbf{81.62}& \textbf{77.1}& \textbf{80.54}& \textbf{75.52}& \textbf{74.87}\\ \cline{2-7}
\multicolumn{1}{c|}{\textbf{supservised}} & \multicolumn{6}{c}{63.46}                                                                                                                                                                          \\ \toprule[1pt]
\multicolumn{7}{l}{10\% SVHN}                                                                                                                                                                                                                                          \\ \hline
\multicolumn{1}{c|}{\textbf{add}}&\multicolumn{1}{c}{63.24}& 59.88&63.35&90.62& 90.63&90.7\\
\multicolumn{1}{c|}{\textbf{concatenate}}&\multicolumn{1}{c}{63.7}& 60.61&63.91&90.85&90.91&90.9\\
\multicolumn{1}{c|}{\textbf{ours}}&\multicolumn{1}{c}{\underline{73.14}}&\underline{72.9}&\underline{74.59}&\underline{91.49}&\textbf{91.59}&\underline{91.44}\\ \cline{2-7}
\multicolumn{1}{c|}{\textbf{supservised}} & \multicolumn{6}{c}{{\underline{\textbf{91.58}}}}                                                                                                                                                                                  \\ \toprule
\multicolumn{7}{l}{10\% CIFAR100}                                                                                                                                                                                                                                      \\ \hline
\multicolumn{1}{c|}{\textbf{add}}& 46.3&\multicolumn{1}{c}{42.18}&\multicolumn{1}{c}{\underline{46.34}}&\multicolumn{1}{c}{39.4}&\multicolumn{1}{c}{\underline{45.2}}&\multicolumn{1}{c}{35.45}          \\
\multicolumn{1}{c|}{\textbf{concatenate}} &\underline{46.46}&\multicolumn{1}{c}{\underline{42.38}}&\multicolumn{1}{c}{46.33}&\multicolumn{1}{c}{\underline{39.67}}&\multicolumn{1}{c}{45.1}& \multicolumn{1}{c}{\underline{35.71}}\\
\multicolumn{1}{c|}{\textbf{ours}}& \textbf{46.64}&\multicolumn{1}{c}{\textbf{45.76}}&\multicolumn{1}{c}{\textbf{46.93}}&\multicolumn{1}{c}{\textbf{42.7}}&\multicolumn{1}{c}{\textbf{46.08}} & \multicolumn{1}{c}{\textbf{41.89}} \\ \cline{2-7}
\multicolumn{1}{c|}{\textbf{supservised}} & \multicolumn{6}{c}{26.51}                                                                                                                                                                                                 \\ \toprule
\end{tabular}}
\label{444}
\end{table*}

\subsubsection{Pretext and Pre-trained model}
For the first stage of SSL, we take ResNet50 as the backbone and learn the semantic representation with MoCo.v2 (pretext task).
For learning the CIFAR10, CIFAR100 and SVHN (and TinyImagenet) representation, we use mini-batch of 256 (64 for TinyImagenet), a dictionary number of 8192, and an initial learning rate of 0.1. We then train for 400 epochs with learning rate deweighted 0.1 at 200 and 300 epochs.
\subsubsection{Downstream tasks}
For the downstream classification, we leverage 4 datasets, natural images(CIFAR10 and CIFAR100) and numbers and letters (SVHN and the Chars74k Englishlmg subset).
We trained downstream networks with the standard cross-entropy loss in the end-to-end way, where the mini-batch size is 128, the SGD weight decay is 0.0001, the SGD momentum is 0.9, and the initial learning rate is 0.1.
We transfer the penultimate layer features of the pre-trained models.
We further train for 200 (100) epochs for natural (numbers and letters) images datasets with learning rate deweighted 0.1 at 100 (50) and 150 (75) epochs.
\subsubsection{Supervised training}
For the supervised paradigm baseline, we take ResNet50 as the backbone and train it with the standard cross-entropy loss in the end-to-end way for 500 epochs, where the mini-batch size is 128, the SGD weight decay is 0.0001, the SGD momentum is 0.9, and the initial learning rate is 0.1 and learning rate deweighted 0.1 at 250 and 375 epochs.

\subsection{Comparison Results}
\subsubsection{Transferring with single pre-trained model}
We first study the performance of transferring knowledge from a single pre-trained model to downstream tasks.
We transfer the representation to downstream tasks by training a task-specific linear classifier (w/o SA).
Furthermore, to validate the effectiveness of the self-attention block, we train an additional self-attention block (w/ SA) with a linear classifier.
The results are shown in Table-\ref{single pre-trained model}, where we draw the following two conclusions.

\begin{itemize}
\item \textit{When the specific downstream task is dissimilar with the pretext, simply training a task-specific classifier will significantly drop the performance (e.g. transferring from CIFAR10 to SVHN).}
\item Self-attention blocks can effectively extract downstream task-related knowledge, which improves the performance of downstream tasks.
\end{itemize}

\subsubsection{Transferring with the representation bank}

In this subsection, we conduct experiments on transferring representation from the bank by (a) adding the representations (add), (b) concatenating the pre-trained representation (concatenate), (c) our proposed attention-based representation aggregation (ours).
Table-III summarizes the experimental results on two different categories (natural image and numbers/letters) downstream datasets. Table-IV  summarizes the result on downstream tasks with small amounts of labeled data (10\%).
We conclude by following several observations.

\begin{table}[h]
\scriptsize
\centering
\caption{Comprehensive Study of Attention Blocks in the framework.}
\scriptsize
	\label{exp:1}
\renewcommand\arraystretch{1.7}
	\setlength{\tabcolsep}{1.8mm}{
\begin{tabular}{c|ccccc}
\bottomrule
\hline\multirow{2}{*}{ Pre-trained models } & \multicolumn{5}{c}{ Attention Blocks Types } \\
\cline { 2 - 6 } & SA only & CA only & SA2CA & CA2SA & SCA \\
\hline CIFAR10+CIFAR100 & $72.84$ & $72.11$ & $\mathbf{7 3 . 4 6}$ & $73.12$ & $72.80$ \\
\hline CIFAR10+Tiny & $72.00$ & $72.52$ & $\mathbf{7 2 . 9 0}$ & $72.70$ & $72.02$ \\
\hline CIFAR100+Tiny & $73.52$ & $74.37$ & $\mathbf{7 4 . 5 9}$ & $74.34$ & $73.79$ \\
\hline SVHN+CIFAR10 & $\mathbf{9 1 . 6 0}$ & $90.89$ & $91.49$ & $91.03$ & $91.39$ \\
\hline SVHN+CIFAR100 & $91.50$ & $91.10$ & $\mathbf{9 1 . 5 9}$ & $91.13$ & $91.51$ \\
\hline SVHN+Tiny & $91.46$ & $91.40$ & $\mathbf{9 1 . 8 3}$ & $91.43$ & $91.51$ \\
\bottomrule
\end{tabular}}
\label{00}
\end{table}

\textbf{Observation 1.} Comparing experiment results in Table-II and Table-III, it strongly verifies that squeezing out representation from the bank (e.g CIFAR10+Tiny to CIFAR10) is better than a single model (CIFAR10/Tiny to CIFAR10).

\textbf{Observation 2.} Table-III and Table-IV results show that our method is superior to other methods. Meanwhile, we observe that the squeeze representation or downstream task performance from knowledge heterogeneous bank (CIFAR100+SVHN to CIFAR100) is almost consistent with that from knowledge heterogeneous bank (CIFAR100+CIFAR10 to CIFAR100), and even exceeds the best single pre-training model in the bank. In the end, promising results demonstrate the powerful ability of our method to adaptively squeeze out representations from the bank.


\textbf{Observation 3.} The results in Table-IV show that when the downstream task has only a small amount of labeled data, SSL significantly surpasses the supervision paradigm. The experimental results again prove that our method is superior to other methods.


In summary, we draw a beneficial conclusion that selectively squeezing out representations from the representation bank can collect more complementary knowledge to assist downstream tasks learning.



To study the performance of our framework on object localization and object recognition, we apply Grad-CAM \cite{selvaraju2017grad} to 4 different networks: (1) supervised training; (2) pre-trained Tiny net; (3) pre-trained CIFAR10 net; (4) pre-trained CIFAR10 and Tiny. All these networks are trained on the downstream CIFAR10 with 10$\%$ images and labels. We show our experimental results by Figure 5 and analyse them as follows:

\begin{itemize}
	\item {\bf}Image 1-2 shows that the supervised network focuses on the background classes, and our approach effectively overcomes this problem.
	
	\item {\bf}Image 3-12 shows that our network is able to focus on object localization information more precisely and is more robust to irrelevant background information.

   \item {\bf}The CIAFR10+Tiny is better than Tiny and CIFAR10 which means that the multiple pre-trained models are better than the single pre-trained model.
	
\end{itemize}

\begin{figure*}[t!]
	\centering
	\includegraphics[width=0.95\textwidth]{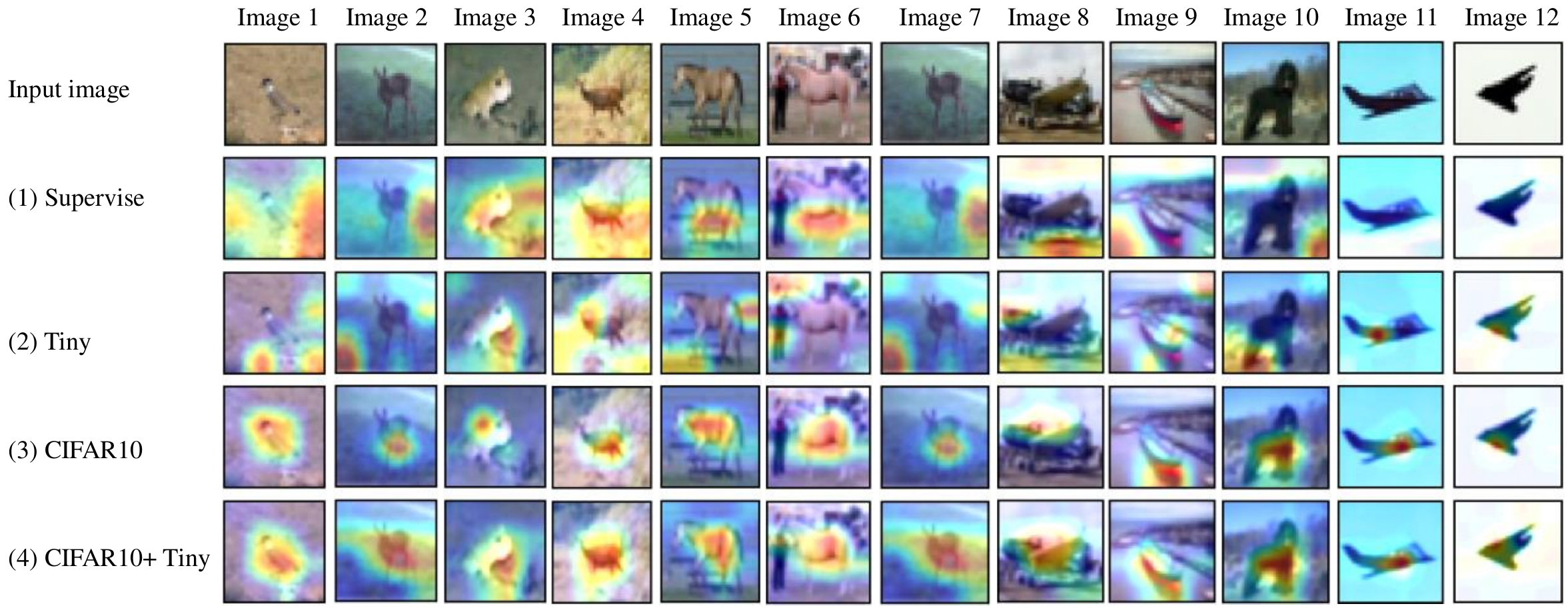}
	\caption{Network visualization with Grad-CAM.
	}
	\label{fig:framework}
\end{figure*}

\begin{figure}[ht!]
\centering
\subfloat{\includegraphics[width=3.6in]{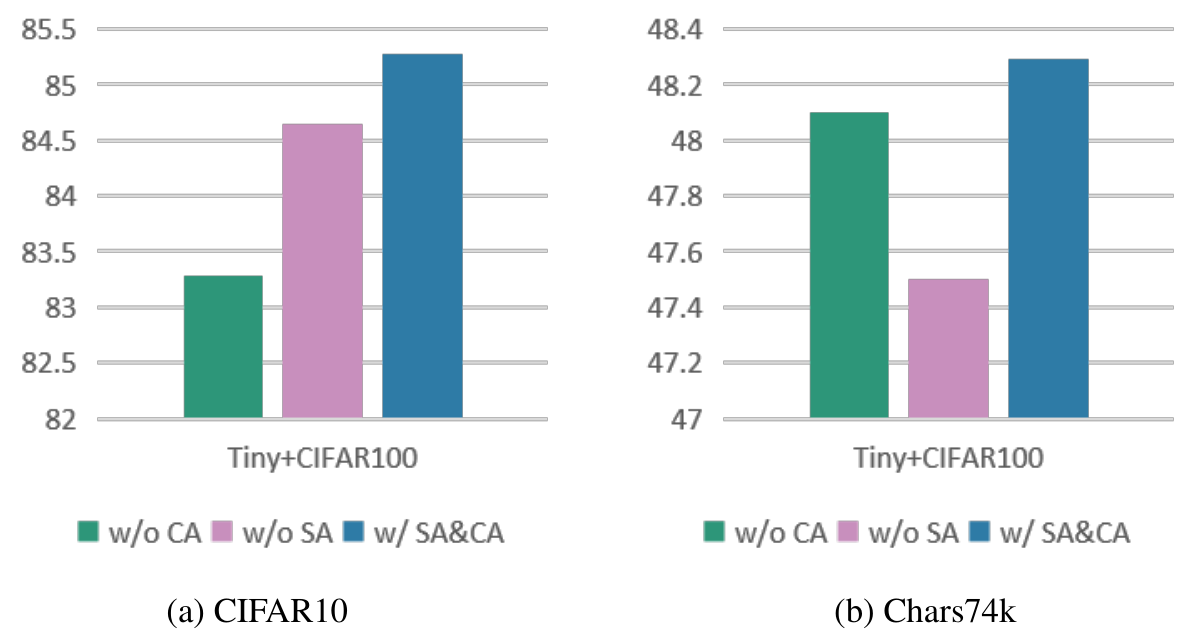}%
\label{7}}
\hfil
\caption{Comparison of the utilities of SA Blocks and CA Blocks separately. Taking models pre-trained on CIFAR100 and TinyImagenet as the source models.}
\label{fig_sim}
\end{figure}

\subsection{Ablation Study}
In this section, we further conduct ablation studies to validate the effectiveness of each component of the proposed model.

\subsubsection{Attention block}
To explore the merit of SAB and CAB in the framework, we respectively remove the SAB (w/o SA) and the CAB (w/o CA), observing its performance in downstream tasks (CIFAR100+Tiny to CIFAR10 \& Chars74k).
As shown in Figure 6, removing SAB or CAB will drop the performance.
As for the CAB, it can aggregate complementary representation in different pretexts (Figure 6-(a), w/o SA is better than w/o CA).
However, we conjecture taking the knowledge heterogeneous bank as the source, SAB is able to filter the irrelevant information (Figure 6-(b), w/o SA falls behind w/o CA).

\begin{figure}[ht!]
\centering
\subfloat{\includegraphics[width=3.5in]{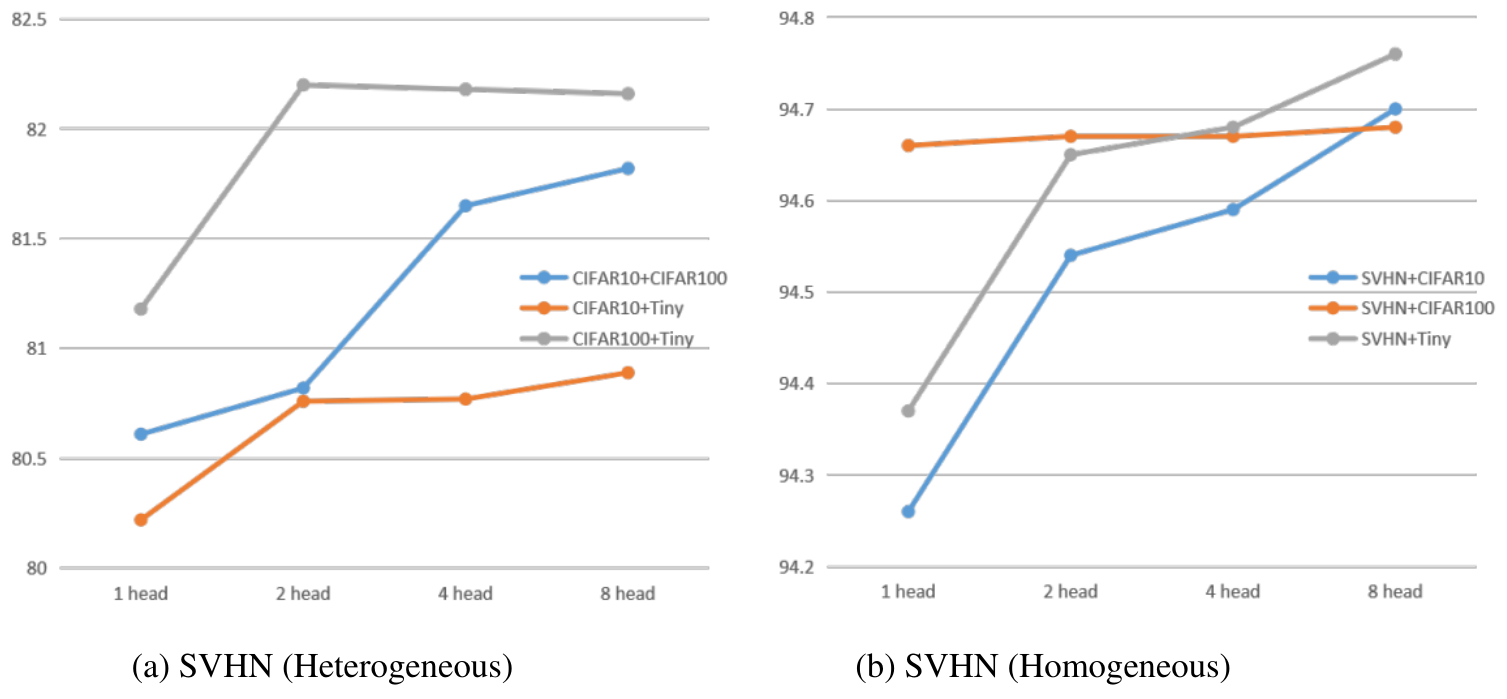}%
\label{5}}
\hfil
\caption{Performance fluctuation figure on changing the number of the attention head in our proposed scheme on downstream task (SVHN). (a) the source pre-trained model did not pre-train on SVHN. In (b), there exist one model in pairs pre-trained on SVHN.}
\label{fig_sim}
\end{figure}

We also conduct extensive comparison study on connection order of both attention blocks to validate the optimality of our architecture. Details are as follows:

we conduct extensive framework comparison experiments with SA Block and CA Block downstream (SVHN with 10$\%$ images and labels). We train 4 different type networks:
(1) using SA block only (SA only), (2) using CA block only (CA only), (3) using SA first and CA later (SA2CA), (4) using CA first and SA later (CA2SA), (5) using SA and CA meanwhile (SCA). The results are shown in the Table-V. From the results, we conclude that using SA first and CA later is the best framework.

\begin{table}[ht!]
\scriptsize
\centering
\caption{Classification accuracy $(\%)$ of downtream task.}
\scriptsize
\renewcommand\arraystretch{1.9}
	\setlength{\tabcolsep}{4.5mm}{
\begin{tabular}{c|c}
\bottomrule
\hline Pre-trained Models & CIFAR10 \\
\hline CIFAR100+Tiny & $85.28$ \\
CIFAR10+CIFAR100 & $86.08$ \\
CIFAR10+Tiny & $86.72$ \\
CIFAR10+CIFAR100+Tiny & $\mathbf{8 6 . 9 8}$ \\\bottomrule
\end{tabular}}
\end{table}
In addition, to improve the representation capacity, as \cite{vaswani2017attention}, we also apply multi-head attention in our experiments. The experimental results are shown in Figure 5. With the increase of the number of heads, the aggregated representation gets stronger representation capacity and performs better on downstream tasks.

\subsubsection{Diverse pretexts}
In order to further  investigate the performance of our model, it is verified by utilizing a more diverse representation bank.
We first apply the multi-head attention mechanism to our framework and construct the multi-head attention (MHA) block. Then, we  investigate the diverse pre-trained models' impact on downstream tasks and conduct comparison experiments of transferring from triple pre-trained models.  Finally, the results in Table-VI  show that adding diverse pre-training knowledge can further improve the learning performance of downstream tasks.The above results once again prove our conclusion that selectively squeezing out representations from the representation bank can collect more complementary knowledge to assist downstream task learning.

\section{Conclusion}
In this paper, we explored how to adaptively squeeze the beneficial knowledge for downstream tasks learning from various pretexts. To this end, we propose a new solution that utilizes the attention mechanism to adaptively squeeze diverse complementary representations from the constructed representation bank to learn downstream tasks. Then, we leverage information theory to prove that the representation/knowledge from diverse pre-trained models can be used as a supplement to downstream tasks. Therefore, we come to a beneficial conclusion that purposefully and strategically squeezing out diverse representation knowledge from the representation bank is helpful to the learning of downstream taskss. For SSL, we provide a novel and effective method for solving how to transfer knowledge from various pretexts. Finally, a large number of experimental results prove that our proposed solution shows the performance of SOTA in collecting various pretexts knowledge and alleviating the negative migration of downstream tasks.

\bibliographystyle{IEEEtran}
\bibliography{mybibfile}
\end{document}

%% file: bare_jrnl_new_sample4.bbl
\begin{thebibliography}{10}
\providecommand{\url}[1]{#1}
\csname url@samestyle\endcsname





\providecommand{\BIBforeignlanguage}[2]{{%
\expandafter\ifx\csname l@#1\endcsname\relax
\typeout{** WARNING: IEEEtran.bst: No hyphenation pattern has been}%
\typeout{** loaded for the language `#1'. Using the pattern for}%
\typeout{** the default language instead.}%
\else
\language=\csname l@#1\endcsname
\fi
#2}}
\providecommand{\BIBdecl}{\relax}
\BIBdecl

\bibitem{wang2020revisiting}
Y.~Wang, Z.~Ni, S.~Song, L.~Yang, and G.~Huang, ``Revisiting locally supervised
  learning: an alternative to end-to-end training,'' in \emph{International
  Conference on Learning Representations}, 2020.

\bibitem{qi2020small}
G.-J. Qi and J.~Luo, ``Small data challenges in big data era: A survey of
  recent progress on unsupervised and semi-supervised methods,'' \emph{IEEE
  Transactions on Pattern Analysis and Machine Intelligence}, 2020.

\bibitem{gidaris2018unsupervised}
S.~Gidaris, P.~Singh, and N.~Komodakis, ``Unsupervised representation learning
  by predicting image rotations,'' \emph{arXiv preprint arXiv:1803.07728},
  2018.

\bibitem{robinson2020contrastive}
J.~Robinson, C.-Y. Chuang, S.~Sra, and S.~Jegelka, ``Contrastive learning with
  hard negative samples,'' \emph{arXiv preprint arXiv:2010.04592}, 2020.

\bibitem{he2020momentum}
K.~He, H.~Fan, Y.~Wu, S.~Xie, and R.~Girshick, ``Momentum contrast for
  unsupervised visual representation learning,'' in \emph{Proceedings of the
  IEEE/CVF Conference on Computer Vision and Pattern Recognition}, 2020, pp.
  9729--9738.

\bibitem{chen2020simple}
T.~Chen, S.~Kornblith, M.~Norouzi, and G.~Hinton, ``A simple framework for
  contrastive learning of visual representations,'' \emph{arXiv preprint
  arXiv:2002.05709}, 2020.

\bibitem{renggli2020model}
C.~Renggli, A.~S. Pinto, L.~Rimanic, J.~Puigcerver, C.~Riquelme, C.~Zhang, and
  M.~Lucic, ``Which model to transfer? finding the needle in the growing
  haystack,'' \emph{arXiv preprint arXiv:2010.06402}, 2020.

\bibitem{shu2021zoo}
Y.~Shu, Z.~Kou, Z.~Cao, J.~Wang, and M.~Long, ``Zoo-tuning: Adaptive transfer
  from a zoo of models,'' in \emph{International Conference on Machine
  Learning}.\hskip 1em plus 0.5em minus 0.4em\relax PMLR, 2021, pp. 9626--9637.

\bibitem{caron2018deep}
M.~Caron, P.~Bojanowski, A.~Joulin, and M.~Douze, ``Deep clustering for
  unsupervised learning of visual features,'' in \emph{Proceedings of the
  European Conference on Computer Vision (ECCV)}, 2018, pp. 132--149.

\bibitem{grill2020bootstrap}
J.-B. Grill, F.~Strub, F.~Altch{\'e}, C.~Tallec, P.~Richemond, E.~Buchatskaya,
  C.~Doersch, B.~Avila~Pires, Z.~Guo, M.~Gheshlaghi~Azar \emph{et~al.},
  ``Bootstrap your own latent-a new approach to self-supervised learning,''
  \emph{Advances in Neural Information Processing Systems}, vol.~33, 2020.

\bibitem{ericsson2020well}
L.~Ericsson, H.~Gouk, and T.~M. Hospedales, ``How well do self-supervised
  models transfer?'' \emph{arXiv preprint arXiv:2011.13377}, 2020.

\bibitem{nguyen2020leep}
C.~V. Nguyen, T.~Hassner, C.~Archambeau, and M.~Seeger, ``Leep: A new measure
  to evaluate transferability of learned representations,'' \emph{arXiv
  preprint arXiv:2002.12462}, 2020.

\bibitem{ngiam2011multimodal}
J.~Ngiam, A.~Khosla, M.~Kim, J.~Nam, H.~Lee, and A.~Y. Ng, ``Multimodal deep
  learning,'' in \emph{ICML}, 2011.

\bibitem{srivastava2014multimodal}
N.~Srivastava and R.~Salakhutdinov, ``Multimodal learning with deep boltzmann
  machines,'' \emph{The Journal of Machine Learning Research}, vol.~15, no.~1,
  pp. 2949--2980, 2014.

\bibitem{noroozi2018boosting}
M.~Noroozi, A.~Vinjimoor, P.~Favaro, and H.~Pirsiavash, ``Boosting
  self-supervised learning via knowledge transfer,'' in \emph{Proceedings of
  the IEEE Conference on Computer Vision and Pattern Recognition}, 2018, pp.
  9359--9367.

\bibitem{tishby2015deep}
N.~Tishby and N.~Zaslavsky, ``Deep learning and the information bottleneck
  principle,'' in \emph{2015 IEEE Information Theory Workshop (ITW)}.\hskip 1em
  plus 0.5em minus 0.4em\relax IEEE, 2015, pp. 1--5.

\bibitem{papadopoulos2021hard}
A.~Papadopoulos, P.~Korus, and N.~Memon, ``Hard-attention for scalable image
  classification,'' \emph{arXiv preprint arXiv:2102.10212}, 2021.

\bibitem{wang2020hard}
D.~Wang, A.~Haytham, J.~Pottenburgh, O.~Saeedi, and Y.~Tao, ``Hard attention
  net for automatic retinal vessel segmentation,'' \emph{IEEE Journal of
  Biomedical and Health Informatics}, vol.~24, no.~12, pp. 3384--3396, 2020.

\bibitem{shankar2018surprisingly}
S.~Shankar, S.~Garg, and S.~Sarawagi, ``Surprisingly easy hard-attention for
  sequence to sequence learning,'' in \emph{Proceedings of the 2018 Conference
  on Empirical Methods in Natural Language Processing}, 2018, pp. 640--645.

\bibitem{jain2020sarcasm}
D.~Jain, A.~Kumar, and G.~Garg, ``Sarcasm detection in mash-up language using
  soft-attention based bi-directional lstm and feature-rich cnn,''
  \emph{Applied Soft Computing}, vol.~91, p. 106198, 2020.

\bibitem{mcclenny2020self}
L.~McClenny and U.~Braga-Neto, ``Self-adaptive physics-informed neural networks
  using a soft attention mechanism,'' \emph{arXiv preprint arXiv:2009.04544},
  2020.

\bibitem{guan2021predicting}
Y.~Guan, H.~Cui, Y.~Xu, Q.~Jin, T.~Feng, H.~Tu, P.~Xuan, W.~Li, L.~Wang, and
  B.-L. Duh, ``Predicting esophageal fistula risks using a multimodal
  self-attention network,'' in \emph{International Conference on Medical Image
  Computing and Computer-Assisted Intervention}.\hskip 1em plus 0.5em minus
  0.4em\relax Springer, 2021, pp. 721--730.

\bibitem{guo2021ssan}
X.~Guo, X.~Guo, and Y.~Lu, ``Ssan: Separable self-attention network for video
  representation learning,'' in \emph{Proceedings of the IEEE/CVF Conference on
  Computer Vision and Pattern Recognition}, 2021, pp. 12\,618--12\,627.

\bibitem{li2021cross}
Z.~Li, F.~Ling, C.~Xu, C.~Zhang, and H.~Ma, ``Cross-media hash retrieval using
  multi-head attention network,'' in \emph{2020 25th International Conference
  on Pattern Recognition (ICPR)}.\hskip 1em plus 0.5em minus 0.4em\relax IEEE,
  2021, pp. 1290--1297.

\bibitem{wang2020cascade}
J.~Wang, X.~Peng, and Y.~Qiao, ``Cascade multi-head attention networks for
  action recognition,'' \emph{Computer Vision and Image Understanding}, vol.
  192, p. 102898, 2020.

\bibitem{shang2021span}
X.~Shang, Q.~Ma, Z.~Lin, J.~Yan, and Z.~Chen, ``A span-based dynamic local
  attention model for sequential sentence classification,'' in
  \emph{Proceedings of the 59th Annual Meeting of the Association for
  Computational Linguistics and the 11th International Joint Conference on
  Natural Language Processing (Volume 2: Short Papers)}, 2021, pp. 198--203.

\bibitem{hanunggul2019impact}
P.~M. Hanunggul and S.~Suyanto, ``The impact of local attention in lstm for
  abstractive text summarization,'' in \emph{2019 International Seminar on
  Research of Information Technology and Intelligent Systems (ISRITI)}.\hskip
  1em plus 0.5em minus 0.4em\relax IEEE, 2019, pp. 54--57.

\bibitem{deng2021superpoint}
S.~Deng, Q.~Dong, B.~Liu, and Z.~Hu, ``Superpoint-guided semi-supervised
  semantic segmentation of 3d point clouds,'' \emph{arXiv preprint
  arXiv:2107.03601}, 2021.

\bibitem{ren2020scga}
D.~Ren, J.~Li, M.~Han, and M.~Shu, ``Scga-net: Skip connections global
  attention network for image restoration,'' in \emph{Computer Graphics Forum},
  vol.~39, no.~7.\hskip 1em plus 0.5em minus 0.4em\relax Wiley Online Library,
  2020, pp. 507--518.

\bibitem{achille2018emergence}
A.~Achille and S.~Soatto, ``Emergence of invariance and disentanglement in deep
  representations,'' \emph{The Journal of Machine Learning Research}, vol.~19,
  no.~1, pp. 1947--1980, 2018.

\bibitem{tishby2000information}
N.~Tishby, F.~C. Pereira, and W.~Bialek, ``The information bottleneck method,''
  \emph{arXiv preprint physics/0004057}, 2000.

\bibitem{vaswani2017attention}
A.~Vaswani, N.~Shazeer, N.~Parmar, J.~Uszkoreit, L.~Jones, A.~N. Gomez,
  {\L}.~Kaiser, and I.~Polosukhin, ``Attention is all you need,'' in
  \emph{Advances in neural information processing systems}, 2017, pp.
  5998--6008.

\bibitem{hu2018squeeze}
J.~Hu, L.~Shen, and G.~Sun, ``Squeeze-and-excitation networks,'' in
  \emph{Proceedings of the IEEE conference on computer vision and pattern
  recognition}, 2018, pp. 7132--7141.

\bibitem{yu2019deep}
Z.~Yu, J.~Yu, Y.~Cui, D.~Tao, and Q.~Tian, ``Deep modular co-attention networks
  for visual question answering,'' in \emph{Proceedings of the IEEE conference
  on computer vision and pattern recognition}, 2019, pp. 6281--6290.

\bibitem{krizhevsky2009learning}
A.~Krizhevsky, G.~Hinton \emph{et~al.}, ``Learning multiple layers of features
  from tiny images,'' 2009.

\bibitem{netzer2011reading}
Y.~Netzer, T.~Wang, A.~Coates, A.~Bissacco, B.~Wu, and A.~Y. Ng, ``Reading
  digits in natural images with unsupervised feature learning,'' 2011.

\bibitem{le2015tiny}
Y.~Le and X.~Yang, ``Tiny imagenet visual recognition challenge,'' \emph{CS
  231N}, vol.~7, 2015.

\bibitem{de2009character}
T.~E. De~Campos, B.~R. Babu, M.~Varma \emph{et~al.}, ``Character recognition in
  natural images.'' \emph{VISAPP (2)}, vol.~7, 2009.

\bibitem{selvaraju2017grad}
R.~R. Selvaraju, M.~Cogswell, A.~Das, R.~Vedantam, D.~Parikh, and D.~Batra,
  ``Grad-cam: Visual explanations from deep networks via gradient-based
  localization,'' in \emph{Proceedings of the IEEE international conference on
  computer vision}, 2017, pp. 618--626.

\end{thebibliography}
